\documentclass[10pt,twocolumn,letterpaper]{article}

\usepackage{cvpr}              
\usepackage{amsmath}
\usepackage{multirow}
\usepackage{booktabs}
\usepackage{wasysym}
\definecolor{cvprblue}{rgb}{0.21,0.49,0.74}
\usepackage[pagebackref,breaklinks,colorlinks,allcolors=cvprblue]{hyperref}
\usepackage[misc]{ifsym}

\def\paperID{45083} 
\def\confName{CVPR}
\def\confYear{2026}

\title{Quantum-Gated Task-interaction Knowledge Distillation for \\ Pre-trained Model-based Class-Incremental Learning}

\author{Linjie Li, Huiyu Xiao, Jiarui Cao, Zhenyu Wu\textsuperscript{(\Letter)}, Yang Ji\\
School of Information and Communication Engineering, BUPT\textsuperscript{\footnotemark[3]}, China\\
Engineering Research Center for Information Network, Ministry of Education, China
}

\begin{document}
\maketitle
\footnotetext[2]{Correspondence to: Zhenyu Wu (shower0512@bupt.edu.cn).}
\footnotetext[3]{BUPT: Beijing University of Posts and Telecommunications.}
\begin{abstract}
Class-incremental learning (CIL) aims to continuously accumulate knowledge from a stream of tasks and construct a unified classifier over all seen classes. Although pretrained models (PTMs) have shown promising performance in CIL, they still struggle with the entanglement of multi-task subspaces, leading to catastrophic forgetting when task routing parameters are poorly calibrated or task-level representations are rigidly fixed. To address this issue, we propose a novel Quantum-Gated Task-interaction Knowledge Distillation (QKD) framework that leverages quantum gating to guide inter-task knowledge transfer. Specifically, we introduce a quantum-gated task modulation gating mechanism to model the relational dependencies among task embedding, dynamically capturing the sample-to-task relevance for both joint training and inference across streaming tasks.  Guided by the quantum gating outputs, we perform task-interaction knowledge distillation guided by these task-embedding-level correlation weights from old to new adapters, enabling the model to bridge the representation gaps between independent task subspaces. Extensive experiments demonstrate that QKD effectively mitigates forgetting and achieves state-of-the-art performance. Code is available at: \url{https://github.com/Frank-lilinjie/CVPR26-QKD}
\end{abstract}
\section{Introduction}
Class-incremental learning (CIL) aims to train a single classifier on a sequence of tasks where new classes arrive over time without access to task identities at test time~\cite{CILsurvey,Continualsurvey}. This task-agnostic setting poses a fundamental challenge: samples from different tasks share a common feature space and often occupy overlapping subspaces, causing new task learning to overwrite previously acquired knowledge. The resulting subspace entanglement leads to catastrophic forgetting~\cite{catastrophic,cataph_2} and biased predictions across tasks, particularly under limited memory budgets~\cite{memo}.

Early CIL methods trained backbones solely on the incremental data stream, relying on rehearsal~\cite{icarl,DER}, regularization~\cite{ewc,lwf}, or network expansion~\cite{DER,memo,Foster}. These approaches require extensive retraining and struggle to scale with increasing tasks. Recent advances in large-scale pretrained models (PTMs) have transformed this landscape~\cite{PTMsurveyCIL}. By freezing the backbone as a universal feature extractor, PTM-based CIL reuses strong and stable representations across all tasks while learning only lightweight adaptation modules. Empirical studies demonstrate that this paradigm achieves superior accuracy, robustness, and training efficiency compared to training from scratch, motivating the widespread adoption of PTM-based parameter-efficient fine-tuning (PEFT) in modern CIL methods.

Within PTM-based CIL, two parameter-efficient families are particularly prominent. Prompt-based methods maintain a prompt pool and retrieve prompts by computing similarity between input features and stored keys~\cite{l2p,dualprompt,codaprompt}. While enabling implicit sharing and specialization, this local-similarity-driven selection becomes noisy when task subspaces overlap or drift, causing prompt mismatches that degrade calibration and aggravate forgetting. Adapter-based methods allocate separate lightweight adapters per task while freezing the PTM backbone, isolating task-specific updates and avoiding parameter conflicts~\cite{EASE,li2025mote}. However, treating adapters as nearly independent subspaces ignores cross-task correlations.  When tasks occupy partially overlapping feature regions, the model has no explicit mechanism to represent cross-task correlation or to decide how multiple historical adapters should jointly influence a new sample. Under task-agnostic evaluation, heuristic routing or naive fusion struggles with entangled subspaces, as the model may confuse representations of different tasks when task identity is unknown in inference.

These observations reveal a fundamental limitation underlying both prompt-based and adapter-based PTM-CIL: routing and fusion lack explicit learned task-interaction mechanisms, relying instead on simple similarity in the original feature space or hand-crafted heuristics. Blind prompt retrieval and isolated adapter subspaces are two manifestations of this gap. As tasks accumulate and subspaces overlap more heavily, three challenges emerge: (i) quantifying relevance between current samples and each task modules, (ii) leveraging relevance to decide from which historical tasks the new task should borrow knowledge during training, and (iii) reusing the same relevance information to combine or select adapters during task-agnostic inference. This motivates the central question of this work: \emph{Can we design a unified, learnable mechanism that dynamically quantifies sample-task relevance and applies it consistently for both training-time knowledge transfer and inference-time adaptive routing?}

To address these challenges, we propose Quantum-Gated Task-Interaction Knowledge Distillation (QKD) for PTM-based class-incremental learning. We freeze the pretrained backbone and allocate a lightweight adapter per task to maintain parameter efficiency and prevent interference in the shared feature extractor.
To quantify complex sample-to-task relevance dynamically, we construct compact task embeddings from each adapter and introduce a learnable quantum gating module to capture the geometric mutual information between incoming samples and pervious task embeddings. This gate takes the current sample feature and all task embeddings as input, then maps them into a higher-dimensional Hilbert space via a parameterized quantum circuit~\cite{quantum,nielsen2010quantum}. Quantum superposition and interference naturally encode complex many-way dependencies among overlapping task subspaces in a compact form~\cite{QRKD}. We then measure the quantum states and transform the outcomes into task-interaction scores, which are normalized via softmax to produce sample-to-task relevance coefficients.
To guide training time cross-task knowledge transfer, we employ these sample-to-task relevance coefficients as weights in a correlation-weighted feature distillation loss from historical adapters to the current adapter. This encourages selective absorption of knowledge from highly relevant tasks while suppressing interference from weakly related ones. Considering alignment between training and inference, we reuse the same quantum-derived sample-to-task relevance coefficients at test time for adaptive adapter fusion under task-agnostic evaluation.

We evaluate QKD on five exemplar-free CIL benchmarks: CIFAR-100~\cite{CIFAR}, CUB-200~\cite{CUB}, ImageNet-A~\cite{ImageNetA}, ImageNet-R~\cite{ImageNetR}, and VTAB~\cite{VTAB}. QKD consistently improves both final and average incremental accuracy while maintaining parameter efficiency. Ablations show that replacing the quantum circuit with cosine similarity or neural networks consistently degrades performance, demonstrating that quantum gating provides essential expressive capacity for modeling complex task interactions.
Our contributions are threefold:
\begin{itemize}[leftmargin=*]
    \item To the best of our knowledge, we propose the first quantum computing-based routing framework to CIL. Our quantum circuit exploits overlap and interference to compactly encode complex many-way task dependencies, dynamically quantifying sample-to-task relevance for both training-time knowledge transfer and inference-time adaptive routing.
    \item We design a relevance-guided distillation strategy that selectively transfers knowledge from related adapters while suppressing interference from unrelated ones.
    \item  QKD achieves state-of-the-art or competitive performance across five diverse benchmarks, with ablations confirming that quantum gating consistently outperforms classical controllers in capturing intricate task dependencies.  \looseness=-1
\end{itemize}
\section{Related Work}
\label{sec:related}
\subsection{Class-Incremental Learning (CIL)}
CIL requires deep learning systems to acquire knowledge of new tasks while retaining knowledge of previous ones, with catastrophic forgetting being a central challenge~\cite{catastrophic,CILsurvey,Continualsurvey}. Early CIL methods address this through three main strategies. \textit{Replay-based} methods mitigate forgetting by storing and replaying a subset of previous data. Rebuffi et al.~\cite{icarl} pioneered exemplar preservation in CIL. Liu et al.~\cite{liu2020mnemonics} introduced bi-level optimization to distill new class data into exemplars. Wang et al.~\cite{wang2022memory} balanced exemplar quality and quantity through JPEG-based image compression. 
\textit{Dynamic-network-based} methods expand network capacity for new tasks while freezing old parameters to preserve knowledge. Yan et al. extended the backbone for each new task and aggregated features at the classifier level~\cite{DER}, Zhou et al. decoupled intermediate layers to reduce memory overhead~\cite{memo}, and Li et al. explored parameter-based expansion~\cite{li2024tae}. \textit{Regularization-based} methods balance new and old knowledge by constraining output logits~\cite{lwf,wu2019large}, intermediate features~\cite{hou2019learning,dhar2019learning}, or inter-class relationships~\cite{gao2022r,dong2021few} via penalty terms or knowledge distillation.

\subsection{Class-Incremental Learning with PTMs}
Recent advances in pretrained models have motivated PTM-based CIL approaches~\cite{Continualsurvey,Pilot,PTMsurveyCIL}, which adapt frozen or partially tuned PTMs to sequential tasks while mitigating forgetting. These methods fall into two families.

\textit{Prompt-based} methods maintain a prompt pool and select instance-specific prompts for task adaptation. Wang et al.~\cite{l2p} introduced learnable prompts with a pretrained Vision Transformer and later extended this with general and expert prompts~\cite{dualprompt}. Smith et al.~\cite{codaprompt} improved selection via attention mechanisms. Wang et al.~\cite{Hide} decomposed the continual learning objective into hierarchical components: within-task prediction, task-identity inference, and task-adaptive prediction.
\textit{Adapter-based} methods introduce lightweight modules or selectively tune parameters. Zhang et al.~\cite{slca} fine-tune the backbone with covariance-based regularization, while McDonnell et al.~\cite{ranpac} project features to high-dimensional space with shared covariance. Zhou et al.~\cite{EASE} generate pseudo-prototypes for old tasks on new adapters via cross-task class-prototypes similarity. Li et al.~\cite{li2025mote} employ mixture-of-experts (MoEs) with peer voting for robust fusion. Zhou et al.~\cite{ADAM} integrate multiple PEFT techniques (prompts, adapters, SSF~\cite{ssf}) into the PTM and freeze both backbone and adapters after first-task training, directly generating prototypes for all classes without additional updates. 
Although prompt-based and adapter-based approaches have made notable progress, they still rely on shallow or heuristic similarity measures, which are insufficient for capturing the nonlinear interactions among task-specific subspaces. To address this limitation, we map representations into the quantum Hilbert space, where the enhanced geometric expressiveness enables more faithful modeling of sample–task relationships. This quantum-based formulation facilitates both task routing and coordinated interaction across tasks.

\subsection{Quantum Machine Learning (QML)}

Quantum computing replaces classical bits with qubits, which represent two-level quantum systems whose pure states correspond to unit vectors in the Hilbert space $\mathbb{C}^{2}$~\cite{quantum,nielsen2010quantum}. QML exploits this by constructing parameterized quantum circuits that embed classical data into high-dimensional Hilbert spaces with rich geometric structure~\cite{QML}. A central challenge is quantifying nonlinear correlations between embedded features, where quantum kernel methods have received substantial attention. Existing approaches can be broadly categorized into global and local kernel measurements. The \textit{Fidelity Quantum Kernel (FQK)}, also termed the embedding kernel or quantum state overlap, embeds inputs into an exponentially large 2$^{n}$-dimensional Hilbert space via an $n$-qubit quantum circuit~\cite{FQK}. While FQKs access highly expressive feature spaces using polynomial resources, they exhibit exponential concentration: as circuit depth and width increase, pairwise fidelities collapse toward a constant due to high-dimensional quantum state geometry. This concentration severely limits effective expressivity and generalization in deep quantum kernel models~\cite{FQK_limit}. 
\textit{Projected Quantum Kernels (PQK)} address this limitation by defining similarity through local observables that extract partial information from subsystems~\cite{PQK}. Rather than computing full-state fidelity, PQKs measure quantities such as Pauli-Z expectations on individual qubits and construct classical feature vectors from these outcomes. PQKs preserve expressive power while avoiding concentration effects, offering faster computation and greater noise robustness without sacrificing representation quality. These properties make PQKs particularly suitable for evaluating geometric correlations between sample features and task representations in hybrid quantum-classical learning systems. 
\section{Preliminaries}
\label{sec:prelim}
\subsection{Class-Incremental Learning}

The objective of CIL is to enable a model to acquire knowledge from continuously evolving data streams that introduce new classes while retaining knowledge of previous ones to build a unified classifier over all observed classes~\cite{CILsurvey}. We consider a sequence of $T$ incremental stages $\{ \mathcal{D}_1,\dots,\mathcal{D}_T \}$, where the $t$-th training stage provides a dataset $\mathcal{D}_t = \{(x_i^t, y_i^t)\}_{i=1}^{n_t}$.
Each input $x_i^t \in \mathbb{R}^{d_x}$ is associated with a label $y_i^t \in \mathcal{Y}_t$, where $\mathcal{Y}_t$ denotes the label set of task $t$. We assume disjoint label sets across tasks, that is, $\mathcal{Y}_t \cap \mathcal{Y}_{t'} = \varnothing$ for any $t \neq t'$.
During training stage $t$, the model is updated using only data from $\mathcal{D}_t$ under the exemplar-free setting, which does not store or replay any samples from previous stages. After completing training stage $t$, the model is evaluated on all classes seen so far, namely on the union $\mathcal{Y}_{1:t} = \mathcal{Y}_1 \cup \dots \cup \mathcal{Y}_t$, using a single classifier$f_t : \mathbb{R}^{d_x} \rightarrow \mathcal{Y}_{1:t}$. The central challenge in this setting is to continuously incorporate new classes without degrading performance on previously learned classes.

\subsection{Class-Incremental Learning with PTMs}

In PTM-based CIL, a pretrained Vision Transformer (ViT) is commonly adopted as the backbone encoder~\cite{ViT}. To preserve its generalization capability, the backbone is kept frozen, and PEFT techniques are introduced to adapt the model to downstream tasks~\cite{PTMsurveyCIL}. In this work, we employ lightweight adapter modules for task-specific tuning~\cite{adaptformer}. These compact modules encode task-specific information with a small number of parameters, without compromising the stability of the pretrained backbone.

Suppose the ViT encoder contains $L$ Transformer blocks, each consisting of a self-attention module followed by an MLP layer. For each incremental task $t$, we attach an adapter branch in parallel with the MLP of each block and add its output to the MLP output through a residual connection. Concretely, an adapter consists of a down-projection linear layer, a ReLU activation, and an up-projection linear layer. Given the input feature $x_{\text{in}} \in \mathbb{R}^{d}$ to an MLP layer, the adapted output is
\begin{equation}
    x_{\text{out}}
    =
    \mathrm{MLP}(x_{\text{in}})
    +
    \mathrm{ReLU}\bigl(x_{\text{in}} W_{\text{down}}^{(t)}\bigr) W_{\text{up}}^{(t)},
    \label{eq:adapter}
\end{equation}
where $W_{\text{down}}^{(t)} \in \mathbb{R}^{d \times r}$ and $W_{\text{up}}^{(t)} \in \mathbb{R}^{r \times d}$ are the adapter parameters for task $t$, and $r \ll d$ is the adapter bottleneck dimension. We denote the set of adapters attached to the $L$ blocks for task $t$ as $\mathcal{A}_t = \{ a_t^1,\dots,a_t^L \}$.
This adapter-based PTM CIL setup serves as the backbone for the quantum-gated task-interaction framework introduced in Section~4.

\section{Methodology}
\label{sec:method}

\begin{figure*}[t]
	\begin{center}
		\includegraphics[width=1\textwidth]{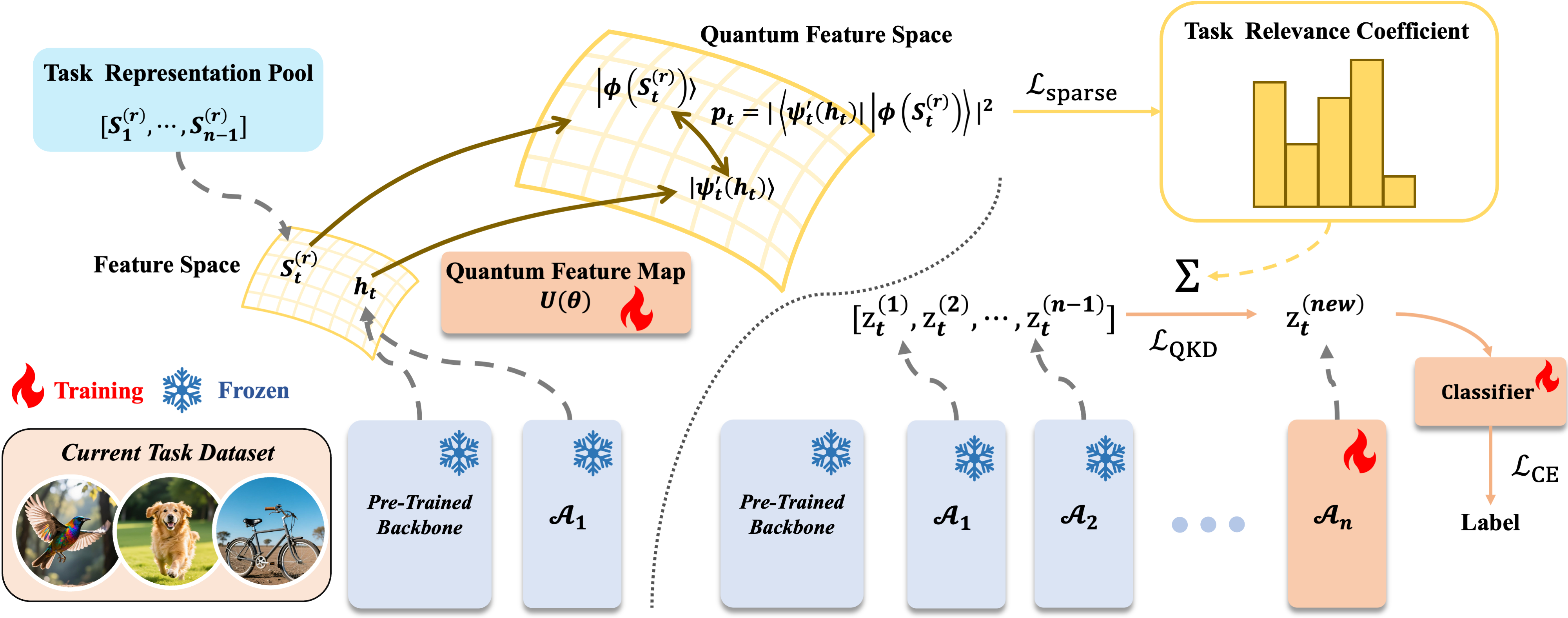}
	\end{center}
	\caption{Illustration of QKD. \textbf{Left:} The current task sample is encoded by the frozen PTM and the first adapter, while old adapters generate task representations forming a task pool. These representations are mapped into the quantum space to compute task correlations. \textbf{Right:} The task correlations produce task scores, which weight the feature-level knowledge distillation to guide new adapter training.} 
    \label{fig: overview}
\end{figure*}

Inspired by the representational advantages of quantum machine learning, we aim to address the subspace entanglement problem in CIL, where features from different tasks overlap in the shared representation space, leading to ambiguity in task identification and degraded knowledge retention. To alleviate this, we introduce a Quantum-Gated Task-interaction Knowledge Distillation (QKD) framework. It consists of a quantum-gated task modulation module (QGTM) that estimates sample–to-task relevance scores (measured by mutual information) via a parameterized quantum circuit, and a task-interaction knowledge distillation (TIKD) scheme that uses these relevance scores to guide cross-task knowledge transfer. The QGTM dynamically produces task-related modulation weights that guide how knowledge should be distilled from historical adapters into the current one.
On top of that, to mitigate inter-task interference and ensure consistent task discrimination, we design a TIKD strategy. This distillation process leverages the quantum-modulated correlation scores as adaptive weights to fuse multi-adapter knowledge at the feature level, promoting better inter-task transfer and subspace separation. Finally, the entire training objective integrates quantum-modulated knowledge distillation and classification constraints into a unified optimization framework, ensuring that the model learns new concepts without overwriting previous knowledge.

\subsection{Quantum-Gated Task Modulation}
\label{sec:QGTM}

As the number of tasks increases in CIL, the model must continuously separate and organize task-specific representations within a shared feature space. Methods relying solely on feature similarity or logit confidence often fail to capture the deeper structural coupling among task subspaces, leading to subspace entanglement in high-dimensional PTM representations. To address this challenge, we propose QGTM, which leverages quantum state encoding, parameterized quantum evolution, and learnable measurement to quantify the geometric mutual information between the incoming sample and previously learned tasks, enabling more accurate task-interaction estimation.

Given an input sample \(x\), we first extract its high-level representation using the frozen ViT backbone together with the adapter set of the first task \(\mathcal{A}_1\):
\begin{equation}
    h_t = f_{\mathrm{ViT}}(x; \mathcal{A}_1).
\end{equation}
We retain \(\mathcal{A}_1\) because it captures generalizable base semantics that benefit subsequent quantum encoding. 

For each previously learned task \(i\), we construct an adapter matrix \(S_i \in \mathbb{R}^{d \times k}\) by stacking the task-specific adapter parameters in \(\mathcal{A}_i\) along the column dimension. We then perform truncated SVD on \(S_i\) and keep the top-\(r\) principal components:
\begin{equation}
    S_i^{(r)} = U_i^{(r)} \Sigma_i^{(r)} V_i^{(r)\top},
\end{equation}
where \(r \ll d\). This truncation preserves the major representation directions of task \(i\), reduces redundancy, and focuses the quantum modulation on the most discriminative components. \looseness=-1

Next we normalize the sample feature and aggregate the dominant task directions into task-level vectors:
\begin{equation}
    \tilde{h}_t = \frac{h_t}{\|h_t\|_2}, \quad
    \tilde{s}_i = \frac{S_i^{(r)} \mathbf{1}}{\|S_i^{(r)} \mathbf{1}\|_2},
\end{equation}
where \(\mathbf{1} \in \mathbb{R}^r\) is an all-one vector used to aggregate the retained principal components, serving as an equal-weight projection that summarizes the dominant subspace of each adapter representation into a single task-level quantum state.  \(\tilde{h}_t\) and \(\tilde{s}_i\) serve as normalized input vectors for subsequent quantum encoding, summarizing the sample representation and the dominant subspace of each task, respectively.

To embed the above vectors into a high-dimensional Hilbert space endowed with expressive geometric structure, we design a shallow, trainable parameterized quantum circuit. The quantum system starts from the computational basis state $|0 \rangle^{\otimes q}$, where $|0 \rangle$ denotes the single-qubit ground state and $q$ means the number of qubits. 

The SVD-reduced sample feature is first mapped into a $q$-dimensional angle vector and encoded via single-qubit rotations:
\begin{equation}
U_{enc}(\tilde{h_{t}}) = \bigotimes_{j=1}^{q}R_{y}(\tilde{h_{t,j}})
\end{equation}
Here, $U_{enc}(\cdot)$ denotes the data-encoding unitary. $\bigotimes$ denotes the tensor product acting on independent qubits. $\tilde{h_{t,j}}$ denotes quit $j$ in the rotation angle of the single-qubit $R_{j}$ gate. This angle encoding transforms the classical feature into amplitude variations on each qubit, enabling nonlinear embedding.

To enhance representational flexibility and make the quantum embedding adaptive across tasks, we introduce learnable rotation parameters $\theta$:
\begin{equation}
U_{var}(\theta) = \bigotimes_{j=1}^{q}R_{y}(\theta_{j})
\end{equation}
Here, $U_{var}(\cdot)$ represents the variational data-encoding unitary. These parameters are updated jointly with the classical components and modulate the quantum embedding dynamically. $\bigotimes$ denotes the tensor product acting on independent qubits. To further capture higher-order cross-dimension dependencies, we apply a CNOT entangling chain:
\begin{equation}
U_{ent} = \prod_{j=1}^{q-1} \mathrm{CNOT} (j\longrightarrow j+1) 
\end{equation}
which establishes quantum correlations among adjacent qubits, and $U_{ent}$ denotes the entangling unitary. Stacking these operations for \(l_q\) layers yields the complete parameterized quantum circuit:
\begin{equation}
    U(\tilde{h}_t; \theta)
    =
    \prod_{l=1}^{l_q}
    \bigl[
        U_{\mathrm{ent}}
        \cdot
        U_{\mathrm{var}}^{(l)}(\theta)
        \cdot
        U_{\mathrm{enc}}^{(l)}(\tilde{h}_t)
    \bigr].
\end{equation}
The resulting quantum state representation is
\begin{equation}
    \bigl|\psi(\tilde{h}_t; \theta)\bigr\rangle
    =
    U(\tilde{h}_t; \theta)\, |0\rangle^{\otimes q},
\end{equation}
where \(|0\rangle^{\otimes q}\) denotes the all-zero computational basis of \(q\) qubits.
With the quantum state $|\psi (\tilde{h}_t;\theta )\rangle$ obtained, we evaluate its relevance to each historical task by measuring the fidelity between the sample state and the task state $|\phi_{i}\rangle = \tilde{s_{i}}$:
\begin{equation}
p_{i} = \left | \left \langle \psi (\tilde{h}_t;\theta ) | \phi _{i} \right \rangle  \right | ^{2}.
\end{equation}
This fidelity score quantifies their geometric proximity in the quantum Hilbert space and serves as a mutual-information-inspired measure of how strongly the current sample interacts with task \(i\).

To further ensure stable knowledge alignment, we introduce a sparsity regularization on the quantum correlation vector \(\alpha = [\alpha_1, \dots, \alpha_{t-1}]\):
\begin{equation}
    \mathcal{L}_{s} = \|\alpha\|_1,
\end{equation}
which encourages the quantum gate to focus only on the most relevant historical adapters, reducing noisy interference from weakly related tasks. \looseness=-1
This design allows the model to maintain plasticity for learning new tasks while preserving stability by selectively distilling useful information from the most correlated past adapters. Through this mechanism, the quantum gating not only serves as a routing module but also becomes a cross-task modulation interface, guiding the optimization direction of inter-task distillation and alleviating subspace entanglement among sequential tasks.

To obtain normalized attention weights reflecting task relevance, we apply a temperature-controlled softmax:
\begin{equation}
\label{eq:temp}
\alpha _{i} = \frac{\mathrm{exp}  (p_{i}/\tau)}{ {\textstyle \sum_{j}\mathrm{exp}  (p_{j}/\tau)} } 
\end{equation}
where $\tau$ is a temperature hyperparameter that controls the sharpness of the modulation.
The resulting $\alpha _{i}$ serves as the sample-to-task relevance coefficient, guiding both the weighted feature fusion and the subsequent cross-task knowledge distillation. In essence, the QGTM models task correlation through learnable quantum rotations and mutual-information-aware projections, allowing the model to adaptively route knowledge across tasks and alleviate feature-space entanglement during training and inference.

\begin{table*}[t]
	\centering
	\resizebox{1.0\textwidth}{!}{%
		\begin{tabular}{@{}lcccccccccc}
			\toprule
			\multicolumn{1}{l}{\multirow{2}{*}{Method}} & 
			\multicolumn{2}{c}{CIFAR B0-Inc10} 
            & \multicolumn{2}{c}{CUB B0-Inc20} 
            & \multicolumn{2}{c}{IN-A B0-Inc10}
			& \multicolumn{2}{c}{IN-R B0-Inc20}
			& \multicolumn{2}{c}{VTAB B0-Inc10} \\
			& $\overline{\mathcal{A}}(\uparrow)$ & $\mathcal{A}_B$($\uparrow$) 
			& $\overline{\mathcal{A}}(\uparrow)$ & $\mathcal{A}_B$($\uparrow$)
			& $\overline{\mathcal{A}}(\uparrow)$ & $\mathcal{A}_B$($\uparrow$)  
			& $\overline{\mathcal{A}}(\uparrow)$ & $\mathcal{A}_B$($\uparrow$)
            & $\overline{\mathcal{A}}(\uparrow)$ & $\mathcal{A}_B$($\uparrow$)
			\\
			\midrule
			Finetune & 79.17 & 67.90 & 69.34 & 52.42 &29.29  &13.17   &72.78 & 60.65 &81.31 & 82.40   \\
            L2P~\cite{l2p}	& 89.14 & 84.18 &78.17 & 65.65 &41.94  &34.76   &76.20 & 69.95 &81.65 & 61.41   \\
            Dualprompt~\cite{dualprompt}	& 88.50 & 84.50 &84.58 & 59.80 &49.98 &35.02 &71.55 & 65.37 &89.32 & 78.65   \\
            CODA-Prompt~\cite{codaprompt}	& 91.44 & 86.70 & 84.69 & 75.15 &47.89 &34.23  &78.23 & 72.62 &87.24 & 75.08  \\
            SimpleCIL~\cite{ADAM}	& 82.37 & 76.27 & 90.56 & 85.20 &60.03 &\underline{49.24}  &61.20 & 54.33 &90.79 & 84.46  \\
            Aper-Finetune~\cite{ADAM}	& 79.25 & 80.98 & 89.94 & 85.24 &61.04 &48.19  &58.62 & 68.23  &91.12 & 85.09 \\
            Aper-VPT-S~\cite{ADAM}	& 85.33 & 82.53 & 91.34 & 84.43 &38.99 &29.03  &62.27 & 68.25  &90.76 & 83.92 \\
            Aper-VPT-D~\cite{ADAM}	& 86.72 & 85.10 &91.00 & 84.52 &44.62 &34.17  &75.36 & 66.78  &91.84 & 85.11 \\
            Aper-Adapter~\cite{ADAM}	& 92.22 & 87.49 &91.65 & 85.94 &59.93 &48.39  & 75.28 & 67.12 &90.80 & 84.46  \\
            Aper-SSF~\cite{ADAM}	& 90.40 & 84.57 & 91.65 & 86.26 &60.98 &47.20  &66.78 & \underline{75.36}  &91.32 & 85.01 \\
            EASE~\cite{EASE}	& 92.14 & 87.67 & 91.02 & 85.50 &61.23 &47.60  &\underline{81.76} & 75.30  &92.19 & 84.58 \\
            MoTE~\cite{li2025mote}	& \underline{93.06} & \underline{88.98} & \underline{91.83} & \underline{86.77} &\underline{63.00} &49.00  &80.44 & 74.88  &\underline{93.56} & \underline{85.94} \\
			\midrule
			QKD(ours) & \bf94.08{\small \textcolor{red}{+1.02}}  & \bf 90.20{\small \textcolor{red}{+1.22}}  & \bf93.54{\small \textcolor{red}{+1.71}} & \bf90.12{\small \textcolor{red}{+3.35}} & \bf65.77{\small \textcolor{red}{+2.77}} & \bf55.56{\small \textcolor{red}{+6.56}}  & \bf81.79{\small \textcolor{red}{+0.03}} & \bf77.67{\small \textcolor{red}{+2.37}} &\bf 95.81{\small \textcolor{red}{+2.25}} & \bf 92.20{\small \textcolor{red}{+6.16}} \\
			\bottomrule
		\end{tabular}
	}
    \vspace{-3mm}
    \caption{\small Average and final accuracy comparison across five datasets. \textbf{Bold} numbers indicate the best performance, while \underline{underlined} values represent the second-best. All methods are evaluated under the exemplar-free setting.}
    \vspace{-3mm}
    \label{tab:benchmark}
\end{table*}

\subsection{Task-Interaction Knowledge Distillation}
\label{sec:TIKD}
Given the quantum-modulated task correlation scores $\alpha _{i}$ obtained from the Eq.~\ref{eq:temp}, we aim to transfer useful knowledge from previously learned adapters to the new one in a relevance-aware manner. Instead of relying on logits-based confidence as in traditional distillation, our approach directly uses the quantum-derived relevance scores to regulate the contribution of each historical adapter, reflecting the mutual information between the current sample and previous task subspaces, and thus implementing a task-interaction-aware knowledge transfer mechanism.
Formally, let \(z_t^{(i)}\) denote the output logits of the \(i\)-th adapter for a given input sample \(x_t\), and \(z_t^{(\mathrm{new})}\) be the logits of the newly introduced adapter for the current task. The quantum-gated knowledge distillation (QKD) loss is defined as
\begin{equation}
    \mathcal{L}_{\mathrm{QKD}}
    =
    \sum_{i=1}^{t-1}
        \alpha_i \cdot
        \mathrm{KL}\bigl(
            \sigma(z_t^{(i)})
            \,\big\|\,
            \sigma(z_t^{(\mathrm{new})})
        \bigr),
\end{equation}
where \(\sigma(\cdot)\) denotes the softmax function.

The quantum-derived weight $\alpha_{i}$ determines how much each previous adapter contributes to the distillation process, enabling adaptive inter-task knowledge transfer based on quantum-correlated relevance.

\subsection{Overall Optimization}
\label{sec:opt}
To jointly optimize the quantum-gated routing and knowledge transfer process, we integrate the classification loss, quantum-gated knowledge distillation loss, and sparsity regularization into a unified training objective. For the current task $t$, given a sample $x_{t}$ and its ground-truth label $y_{t}$, the classification loss is defined as:
\begin{equation}
\label{eq:ce}
\mathcal{L}_{\mathrm{CE} } = -\sum_{c=1}^{C_{t}} y_{t}^{(c)}\mathrm{log} p_{t}^{(c)},
\end{equation}
where $p_{t}^{(c)} = \sigma (\mathbf{z}_{t}^{new} )$ denotes the predicted probability of class $c$ from the new adapter. This term ensures task-specific discriminability and drives the new adapter to learn reliable class representations. The overall optimization objective is therefore formulated as:
\begin{equation}
\label{eq:overall}
\mathcal{L}_{\mathrm{total} } =  \mathcal{L}_{\mathrm{CE} }+\lambda_{\mathrm{kd} }\mathcal{L}_{\mathrm{QKD} }+\lambda_{\mathrm{s} }\mathcal{L}_{\mathrm{s} }
\end{equation}
where $\lambda_{\mathrm{kd}} $ and $\lambda_{\mathrm{s} }$ are the balancing coefficients controlling the relative importance of cross-task knowledge transfer and sparsity regularization. During training, only the current adapter and the quantum gating network are updated, while previously trained adapters remain frozen to prevent forgetting.
During inference, the quantum gate estimates task correlations $\alpha_{i}$ for each input sample and uses them as adaptive weights to combine or select task-specific adapters, ensuring stable performance under task-agnostic testing conditions.  An overview of our algorithm is presented in Fig.~\ref{fig: overview}.

\section{Experiments}
\label{sec:exp}
\subsection{Implementation Details}

\begin{figure*}[t]
	\centering
	\begin{subfigure}{0.3\linewidth}
		\includegraphics[width=1\columnwidth]{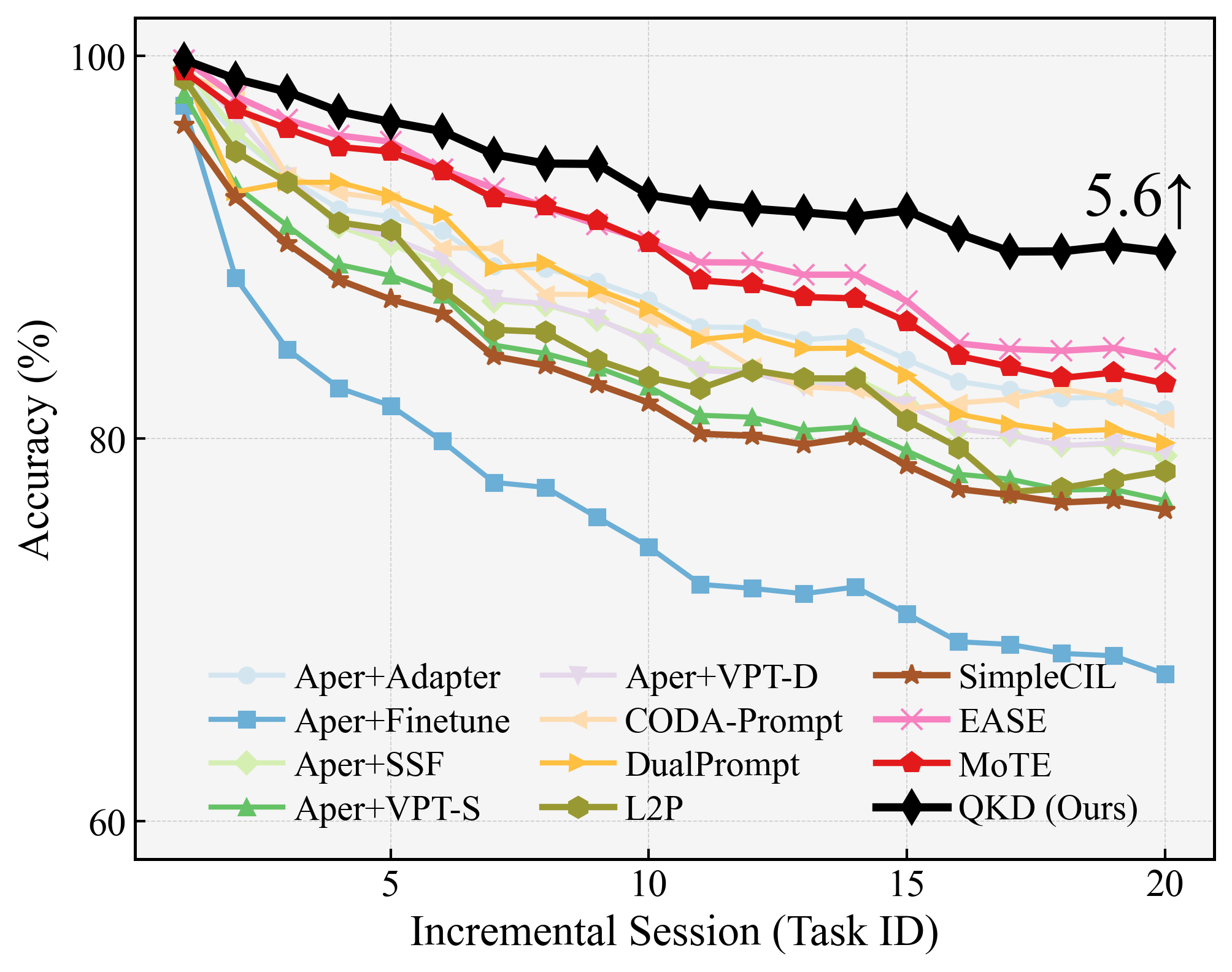}
		\caption{\small CIFAR B0-Inc5}
		\label{fig:benchmark-cifar—5}
	\end{subfigure}
	\hfill
	\begin{subfigure}{0.3\linewidth}
		\includegraphics[width=1\columnwidth]{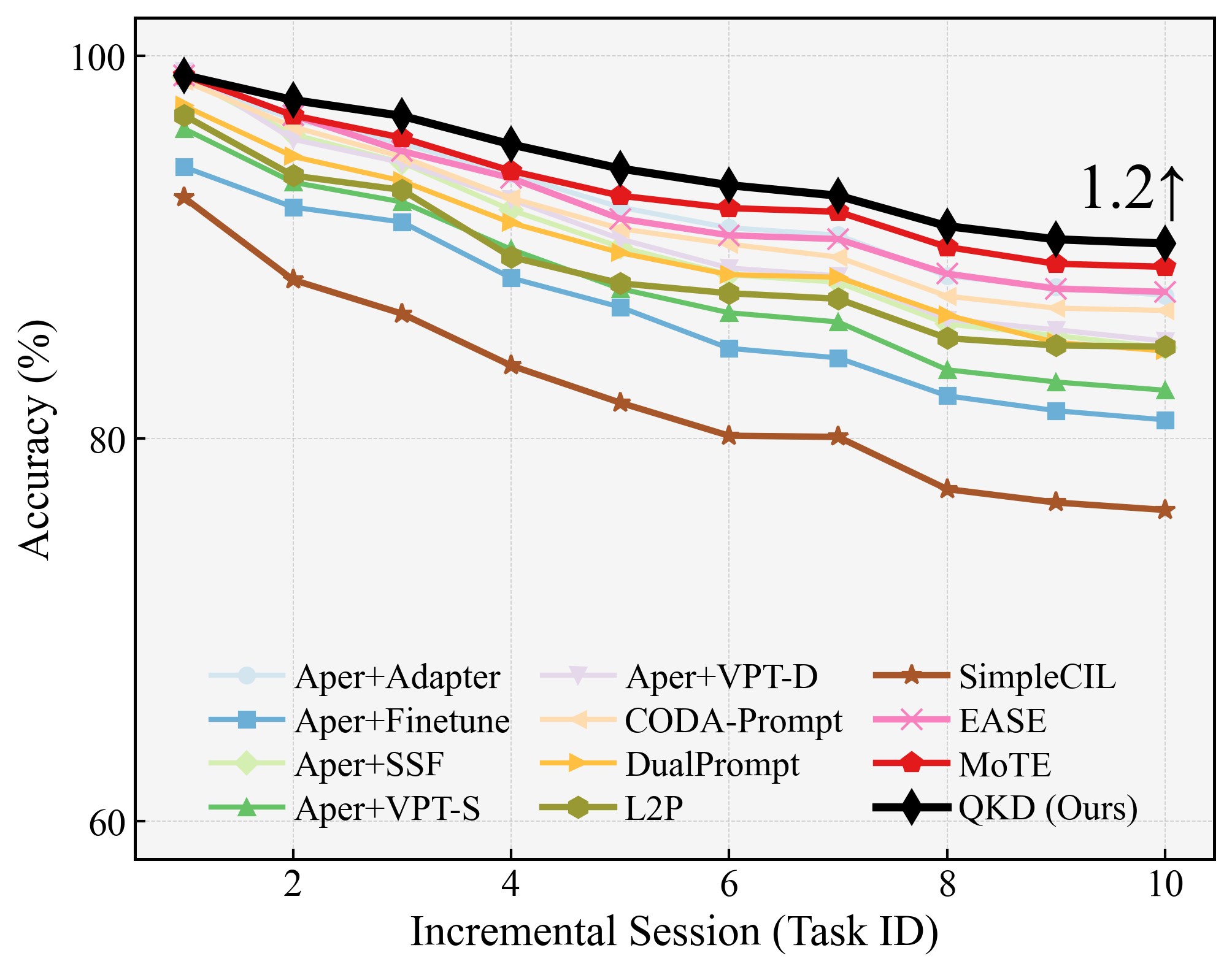}
		\caption{\small CIFAR B0-Inc10}
		\label{fig:benchmark-cifar-10}
	\end{subfigure}
	\hfill
	\begin{subfigure}{0.3\linewidth}
		\includegraphics[width=1\columnwidth]{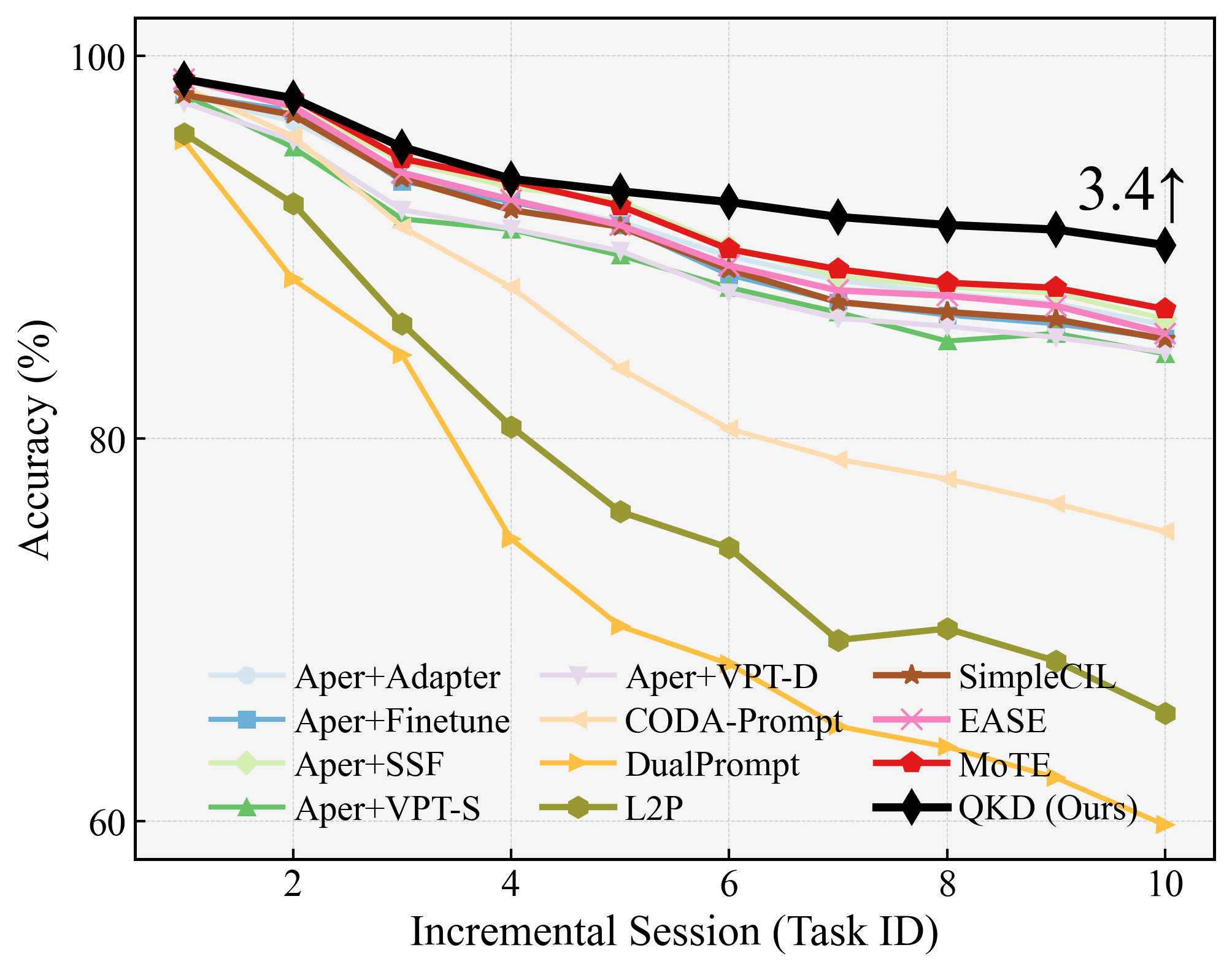}
		\caption{\small CUB B0-Inc20}
		\label{fig:benchmark-cub}
	\end{subfigure}

    \begin{subfigure}{0.3\linewidth}
		\includegraphics[width=1\columnwidth]{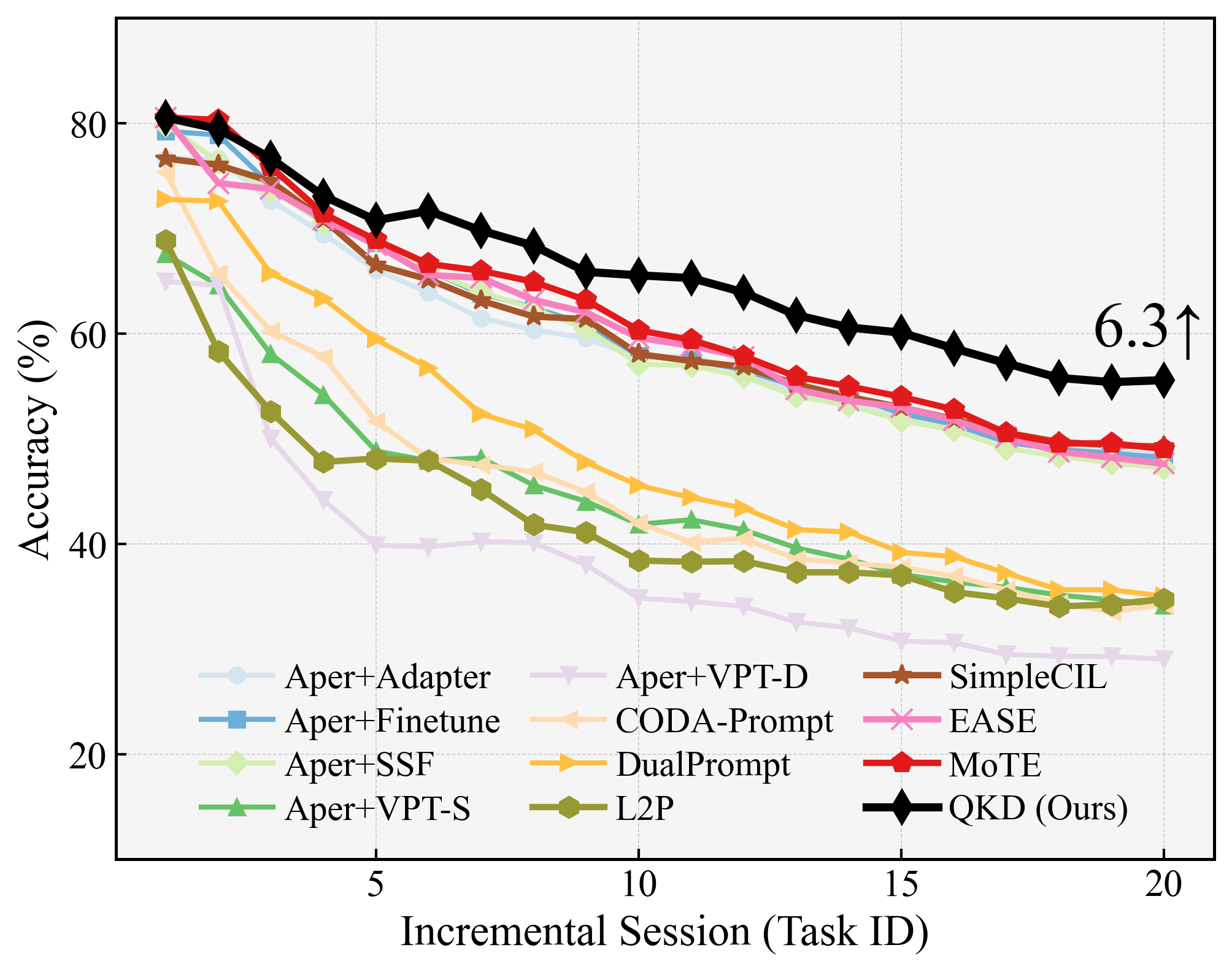}
		\caption{\small IN-A B0-Inc10}
		\label{fig:benchmark-ina}
	\end{subfigure}
	\hfill
	\begin{subfigure}{0.3\linewidth}
		\includegraphics[width=1\columnwidth]{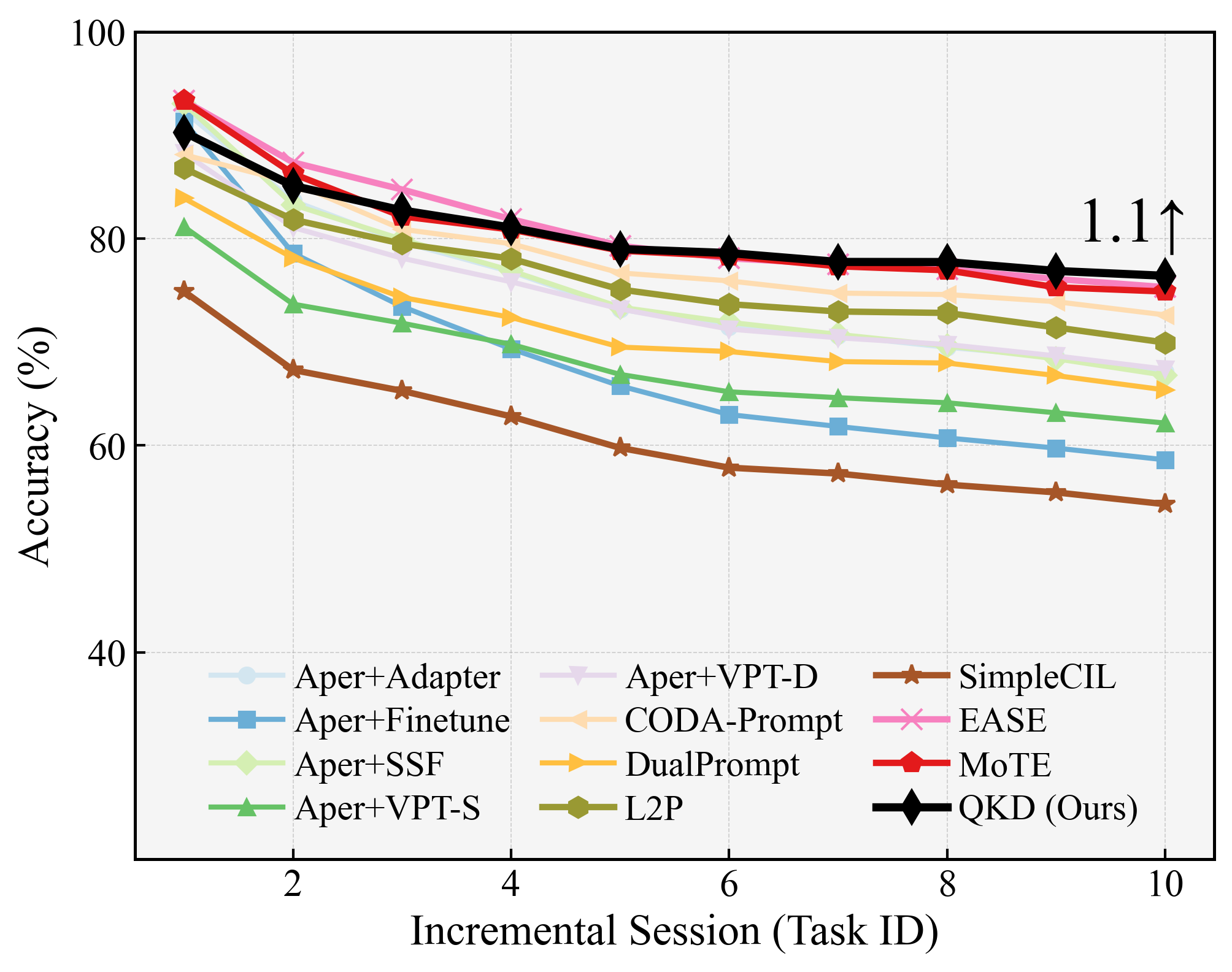}
		\caption{\small IN-R B0-Inc20}
		\label{fig:benchmark-inr}
	\end{subfigure}
	\hfill
	\begin{subfigure}{0.3\linewidth}
		\includegraphics[width=1\columnwidth]{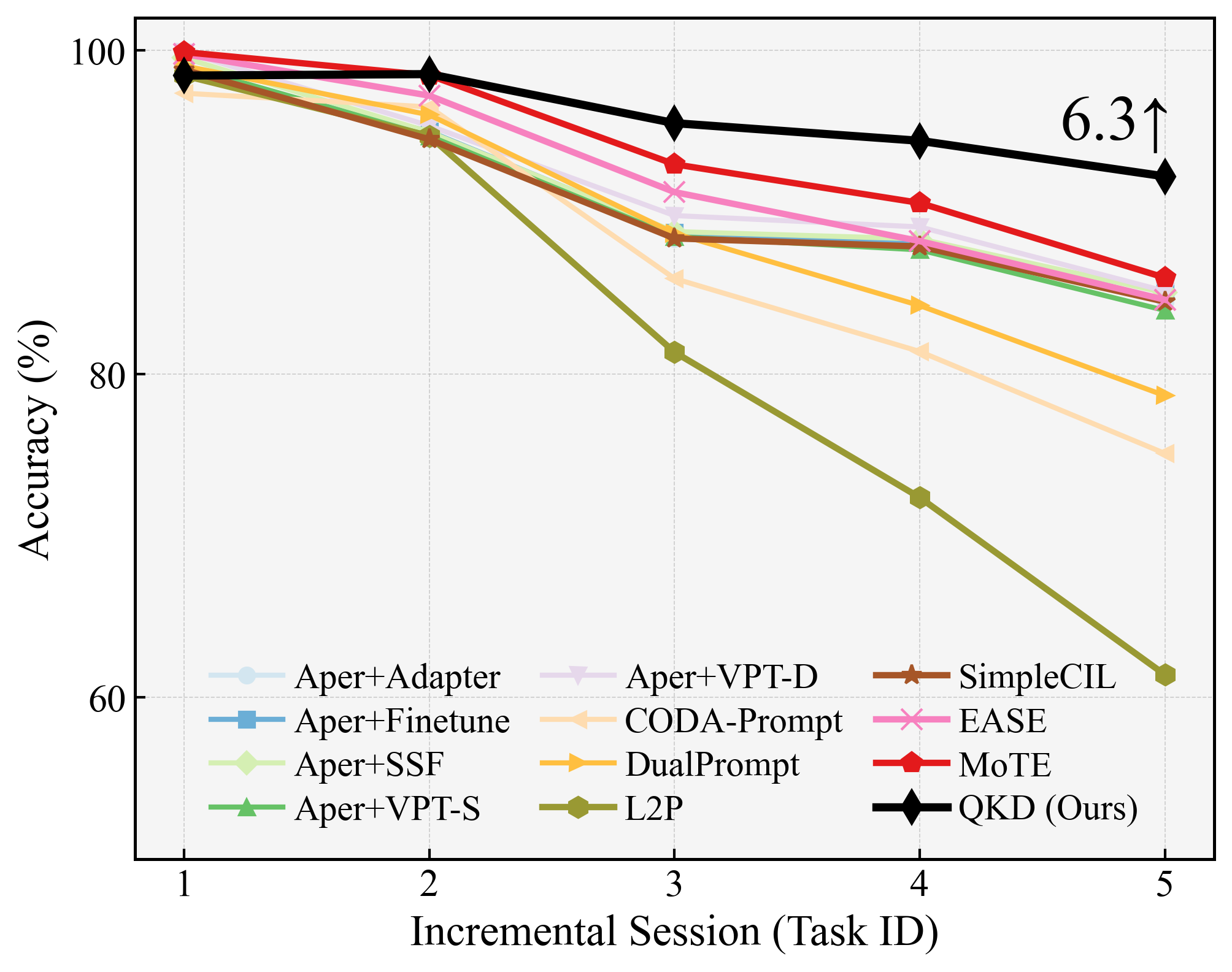}
		\caption{\small VTAB B0-Inc10}
		\label{fig:benchmark-vtab}
	\end{subfigure}
	\caption{Performance curve of various methods under different settings. The relative improvement of QKD over the second-best method at the final incremental stage is highlighted with numerical annotations.}
    \vspace{-3mm}
	\label{fig:benchmarkcompare}
\end{figure*}

\textbf{Datasets:} Given that PTMs encapsulate extensive knowledge from upstream tasks, our evaluation methodology adheres to the protocol established by~\cite{ADAM}. We assess performance on five benchmark datasets: CIFAR100~\cite{CIFAR}, CUB200~\cite{CUB}, ImageNet-R~\cite{ImageNetR}, ImageNet-A~\cite{ImageNetA}, and VTAB~\cite{VTAB}. Specifically, VTAB contains 50 classes, CIFAR100 contains 100 classes, and CUB200, ImageNet-R, and ImageNet-A each contain 200 classes. \textbf{IN-R} denotes ImageNet-R, and \textbf{IN-A} denotes ImageNet-A in this paper. 

\noindent\textbf{Dataset Split:} We adopt the standard benchmark settings from \cite{icarl, l2p}. The class split notation 'B-$m$ Inc-$n$' is used, where $m$ denotes the number of classes in the initial base task, and $n$ represents the number of new classes in each incremental task. 

\noindent\textbf{Training Details:} All models were implemented using PyTorch~\cite{pytorch}. For fair comparison, all methods utilize the same pre-trained models \textbf{ViT-B/16-IN21K}, which is pre-trained on ImageNet21K~\cite{imagenet}. In our method, we use a batch size of 32 and train for 20 epochs with the SGD optimizer. The learning rate is set to 0.05 and follows a cosine annealing decay schedule~\cite{cosine}. The projection dimension $r$ of the adapter is 64, while the SVD dimension is set to 12. In Eq.~\ref{eq:overall}, we set $\lambda_{KD}$ to 1.0 and $\lambda_{s}$ to 0.05 and  $\tau$ to 1.0 in Eq.~\ref{eq:temp}. \looseness=-1

\noindent\textbf{Evaluation Protocol:} We adhere to the standard evaluation benchmark established by~\cite{icarl}. The Top-1 accuracy after the $b$-th incremental stage is denoted as $\mathcal{A}_b$. We report the performance after the final stage, $\mathcal{A}_B$, and the average incremental accuracy, $\overline{\mathcal{A}} = \frac{1}{B} \sum_{b=1}^{B} \mathcal{A}_b$, which represents the mean performance across all completed stages.

\subsection{Benchmark Comparison}

\begin{table}[t]
	\centering
	\resizebox{1.0\columnwidth}{!}{%
		\begin{tabular}{@{}lccccc}
			\toprule
			\multicolumn{1}{l}{\multirow{2}{*}{Method}} & 
            \multicolumn{1}{l}{\multirow{2}{*}{Exemplars}} & 
			\multicolumn{2}{c}{CIFAR B0-Inc10}
			& \multicolumn{2}{c}{IN-R B0-Inc20}\\
            
			& & $\overline{\mathcal{A}}(\uparrow)$ & $\mathcal{A}_B$($\uparrow$) 
			&  $\overline{\mathcal{A}}(\uparrow)$ & $\mathcal{A}_B$($\uparrow$)
			\\
			\midrule
			iCaRL~\cite{icarl} & 20/class & 82.46 & 73.87 & 72.42 & 60.67  \\
            DER~\cite{DER} & 20/class & 86.04 & 77.93 & 80.48 & 74.32  \\
            FOSTER~\cite{Foster} & 20/class & 89.87 & 84.91 & 81.34 & 74.48  \\
            MEMO~\cite{memo} & 20/class & 92.17 & 89.26 & 74.80 & 66.62  \\
			\midrule
			QKD(ours) & \bf 0  & \bf94.08 & \bf90.20 & \bf81.79 & \bf77.67 \\
			\bottomrule
		\end{tabular}
	}
    \vspace{-3mm}
    \caption{\small In contrast to conventional exemplar-based CIL approaches, QKD operates without storing any exemplars.}
    \vspace{-5mm}
    \label{tab:exemplar-based}
\end{table}

\begin{figure}[t]
	\centering
	\begin{subfigure}{0.48\linewidth}
		\includegraphics[width=1\columnwidth]{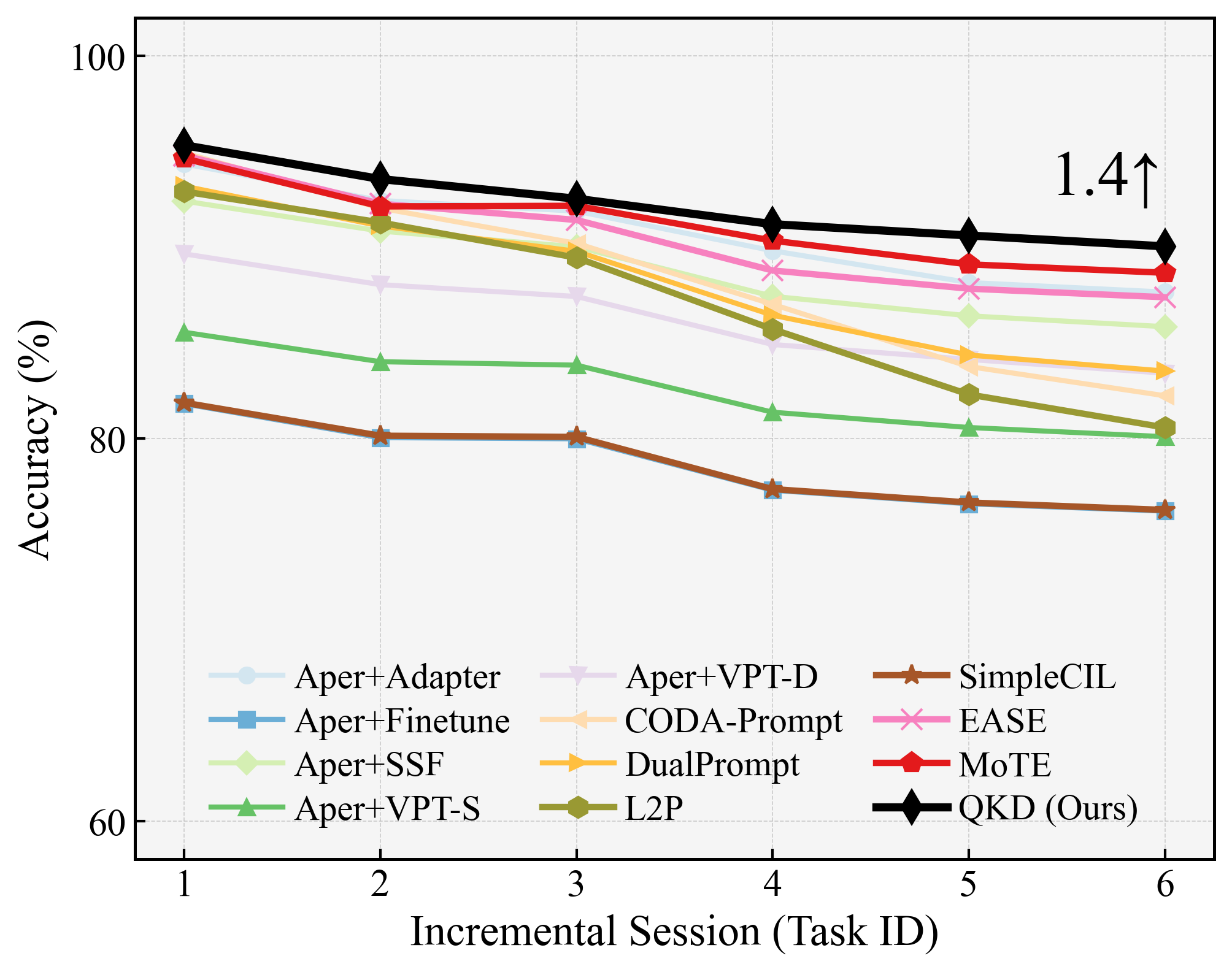}
		\caption{\small CIFAR B50-Inc10}
		\label{fig:benchmark-cifar—50-10}
	\end{subfigure}
	\hfill
	\begin{subfigure}{0.48\linewidth}
		\includegraphics[width=1\columnwidth]{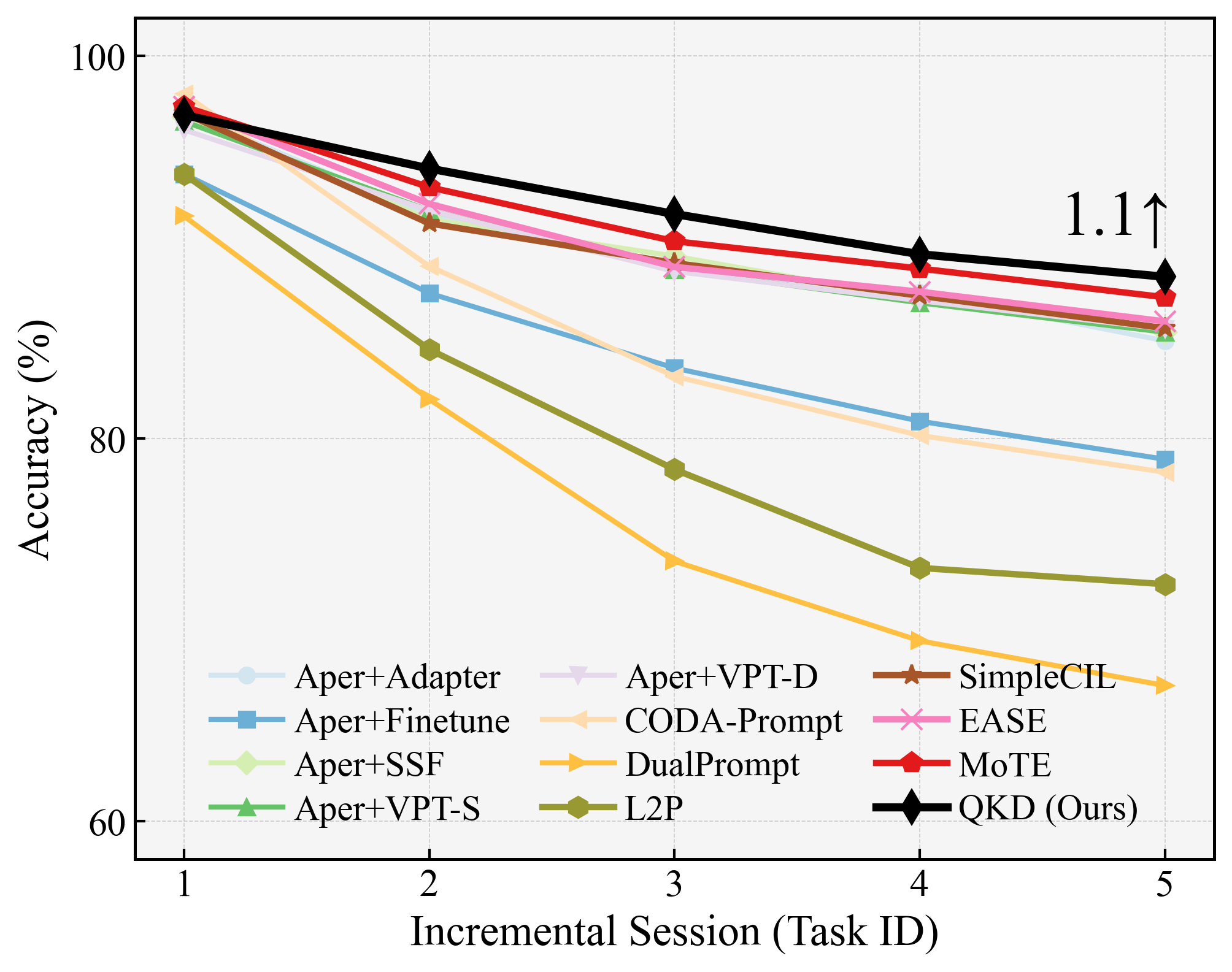}
		\caption{\small CUB B40-Inc40}
		\label{fig:benchmark-CUB-40}
	\end{subfigure}
    \vspace{-3mm}
	\caption{Experimental results with large base classes.}
	\label{fig:large}
    \vspace{-4mm}
\end{figure}

In this section, we conduct a comprehensive experiment to evaluate our method QKD against state-of-the-art baselines across five benchmark datasets. Tab.~\ref{tab:benchmark} reports the accuracy of different methods under the ViT-B/16-IN21K backbone. We observe that QKD consistently achieves the best performance across all evaluation metrics. Furthermore, Fig.~\ref{fig:benchmarkcompare} presents the performance trajectories of various methods across incremental stages. These results collectively demonstrate the effectiveness of our approach.

To further validate the robustness of QKD, we design large-base experiments, as shown in Fig.~\ref{fig:large}. We compare QKD with several SOTA methods. The results show that QKD maintains a clear performance lead. In addition, we compare QKD against traditional CIL methods. As summarized in Tab.~\ref{tab:exemplar-based}, all traditional CIL baselines are equipped with a fixed memory budget of 20 exemplars per class. For fairness, we use the same backbone (ViT-B/16-IN21K) for all baselines. Remarkably, QKD achieves substantially higher performance despite being exemplar-free.

\subsection{Ablation Study}
\begin{table}[t]
    \centering
    \resizebox{1.0\columnwidth}{!}{%
    \begin{tabular}{ccccccccc}
        \toprule
        \multirow{2}{*}{QGTM} & \multirow{2}{*}{$\mathcal{L}_{\mathrm{QKD}}$} & \multirow{2}{*}{$\mathcal{L}_{\mathrm{s}}$} & 
        \multicolumn{2}{c}{CIFAR B0-Inc20} &
        \multicolumn{2}{c}{CUB B0-Inc20} & 
        \multicolumn{2}{c}{IN-A B0-Inc10} \\
        & & &  $\overline{\mathcal{A}}(\uparrow)$ & $\mathcal{A}_B$($\uparrow$) 
		&  $\overline{\mathcal{A}}(\uparrow)$ & $\mathcal{A}_B$($\uparrow$)
        &  $\overline{\mathcal{A}}(\uparrow)$ & $\mathcal{A}_B$($\uparrow$)\\
        \midrule
        $\times$     & $\times$     & $\times$     & 78.42 & 71.29 & 65.26 & 61.98 & 42.83 & 37.89\\
        $\checkmark$ & $\times$     & $\times$     & 88.59 & 87.77 & 91.73 & 88.28 & 62.15 & 51.79\\
        $\checkmark$ & $\checkmark$ & $\times$     & 92.69 & 89.28 & 92.89 & 89.42 & 64.92 & 54.98\\
        $\checkmark$ & $\checkmark$ & $\checkmark$ & \bf93.76 & \bf90.24 & \bf93.45 & \bf90.20 & \bf65.77 & \bf55.65\\
        \bottomrule
    \end{tabular}
    }
    \vspace{-3mm}
    \caption{\small Ablation Study on the QGTM, $\mathcal{L}_{\mathrm{QKD}}$ and $\mathcal{L}_{\mathrm{s}}$ in QKD.}
    \vspace{-2mm}
    \label{tab:abla}
\end{table}
In this section, we conduct an ablation study on the three key components of QKD: the QGTM, TIKD, and task sparsity regularization. As shown in Tab.~\ref{tab:abla}, when all three components are removed, the model falls back to randomly selecting an adapter for inference. This leads to a significant performance drop, which supports our hypothesis that indiscriminately mixing task features causes severe confusion in the shared representation space. Moreover, incorporating each component incrementally brings consistent performance gains, demonstrating that all three modules contribute positively to the overall effectiveness of QKD.

\subsection{Further Analysis}
\begin{table}[t]
\resizebox{1.0\columnwidth}{!}{
\begin{tabular}{llcccccc}
\hline
Benchmark                     & Method    & Params & Time(s) & Mem(GB) & Lat/img & Lat/batch & Avg(\%)        \\ \hline
\multirow{4}{*}{INR B0-Inc20} & MLP       & 1673416             & 300.16           & 2.13            & 13.23             & 136.39            & 80.25          \\
                              & Cosine    & 1343432             & 211.25           & 2.12            & 14.07             & 180.24            & 80.32          \\
                              & Attention & 10798280            & 306.80           & 2.17            & 14.58             & 178.75            & 80.08          \\ \cline{2-8} 
                              & Quantum   & 1349620             & 326.85           & 2.12            & 22.84             & 222.40            & \textbf{81.79} \\ \hline
\multirow{4}{*}{CUB B0-Inc20} & MLP       & 1673416             & 139.49           & 3.43            & 19.48             & 264.77            & 92.63          \\
                              & Cosine    & 1343432             & 114.77           & 3.43            & 13.63             & 340.64            & 92.72          \\
                              & Attention & 10798280            & 142.25           & 3.46            & 12.11             & 272.79            & 92.57          \\ \cline{2-8} 
                              & Quantum   & 1349620             & 156.59           & 3.43            & 20.76             & 425.52            & \textbf{93.54} \\ \hline
\end{tabular}
}
\vspace{-3mm}
\caption{\small Computational Analysis of different task recognizers.
}
\vspace{-2mm}
\label{tab:computation}
\end{table}

\begin{figure}[t]
	\centering
	\begin{subfigure}{0.48\linewidth}
		\includegraphics[width=1\columnwidth]{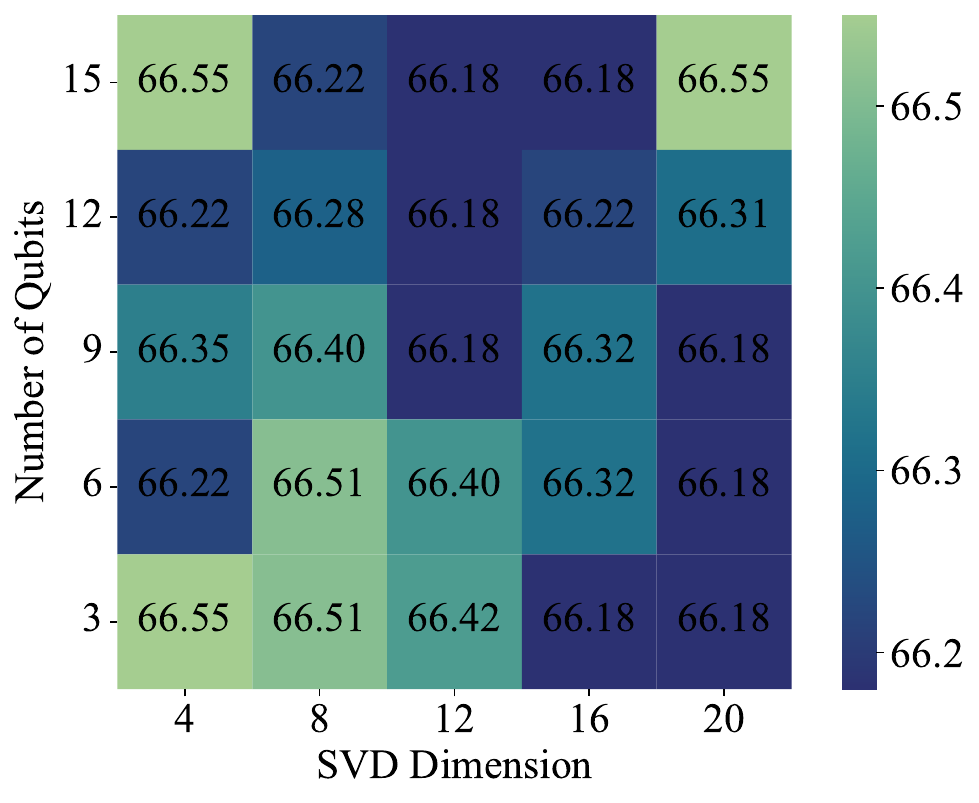}
		\caption{\small IN-A B0-Inc20}
		\label{fig:para-ina—20}
	\end{subfigure}
	\hfill
	\begin{subfigure}{0.48\linewidth}
		\includegraphics[width=1\columnwidth]{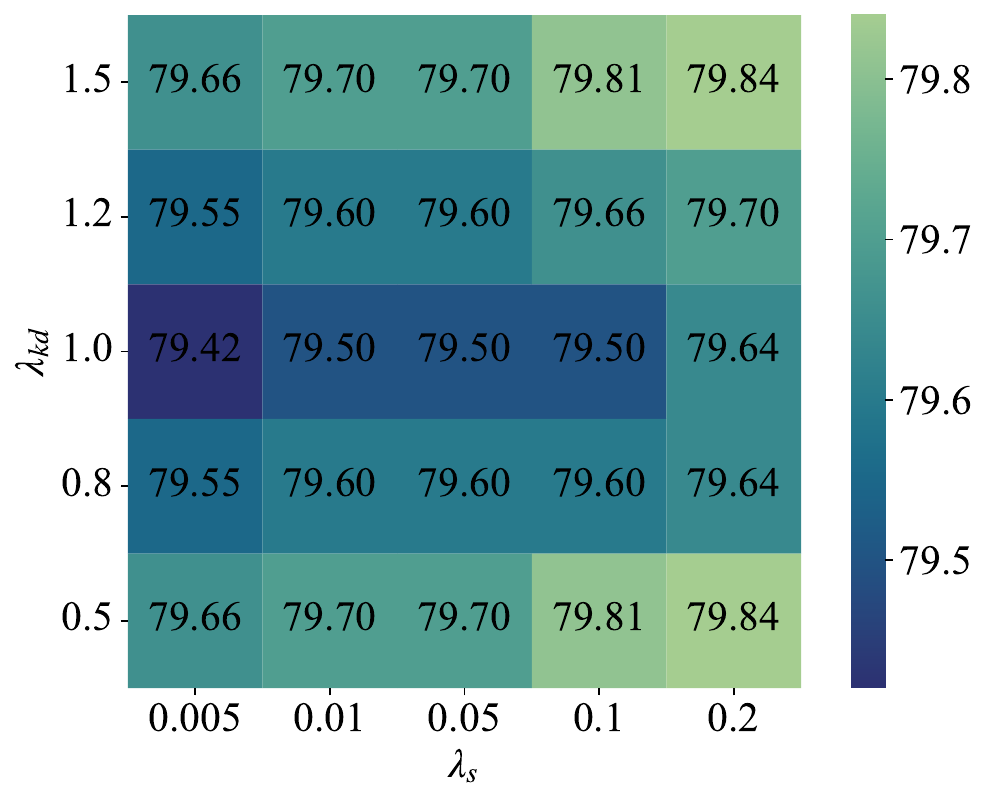}
		\caption{\small IN-R B0-Inc10}
		\label{fig:para-inr—10}
	\end{subfigure}
    \vspace{-3mm}
	\caption{Further analysis on hyperparameter robustness.}
    \vspace{-5mm}
	\label{fig:para}
\end{figure}

\noindent\textbf{Comparison of different task recognizers.} In Tab.~\ref{tab:computation}, we provide a detailed computational comparison.  Owing to the lightweight of quantum gating, its number of trainable parameters, training time, and peak memory consumption are comparable to the classic. While the inference latency is slightly higher, the additional overhead is acceptable in light of the consistent performance gains achieved.

\noindent\textbf{Hyperparameter Robustness.} QKD involves two main groups of hyperparameters: the number of qubits in QGTM and the dimension of the SVD-reduced feature, and $\lambda_{\mathrm{kd}}$ and $\lambda_{\mathrm{s}}$ in Eq.~\ref{eq:overall}. To assess the robustness of our method, we conduct extensive experiments across multiple hyperparameter settings on the IN-A B0-Inc20 and IN-R B0-Inc10 benchmarks. Specifically, the SVD dimension is varied within $\{4,8,12,16,20\}$, and the number of qubits is varied within $\{3,6,9,12,15\}$. For the two coefficients in Eq.~\ref{eq:overall}, we sweep $\lambda_{\mathrm{kd}} = \{ 0.5, 0.8, 1.0, 1.2, 1.5\}$ and $\lambda_{\mathrm{s}}= \{0.005, 0.01, 0.05, 0.1, 0.2 \}$. The results, summarized in Fig.\ref{fig:para}, demonstrate that QKD remains consistently stable across a wide range of hyperparameter configurations, indicating strong robustness. 

\begin{table}[t]
\resizebox{1.0\columnwidth}{!}{
\begin{tabular}{lcccccc}
\hline
\multirow{2}{*}{Method} & \multicolumn{2}{c}{CIFAR B0-Inc10}       & \multicolumn{2}{c}{CUB B0-Inc20}         & \multicolumn{2}{c}{VTAB B0-Inc10}        \\ 
                        & $\overline{\mathcal{A}}(\uparrow)$              & $\mathcal{A}_B$($\uparrow$)             & $\overline{\mathcal{A}}(\uparrow)$              & $\mathcal{A}_B$($\uparrow$)             & $\overline{\mathcal{A}}(\uparrow)$              & $\mathcal{A}_B$($\uparrow$)             \\ \hline
$f_{\mathrm{ViT}}(x)$                 & 93.19          & 88.83          & 91.94          & 86.26          & 95.10          & 90.20          \\
$f_{\mathrm{ViT}}(x; \mathcal{A}_1) $             & \textbf{94.08} & \textbf{90.20} & \textbf{93.54} & 90.12          & \textbf{95.81} & \textbf{92.20} \\
$f_{\mathrm{ViT}}(x;{[}\mathcal{A}_1,\mathcal{A}_2{]})$     & 93.82          & 89.74          & 93.26          & \textbf{90.35} & 95.63          & 91.54          \\ \hline
\end{tabular}
}
\vspace{-3mm}
\caption{\small Performance Comparison of Different Inputs.
}
\vspace{-3mm}
\label{tab:adapter_diff}
\end{table}

\noindent\textbf{Comparison of Different Inputs.} We extract high-level representations using $f_{\mathrm{ViT}}(x; \mathcal{A}_1)$, which are used as inputs to the quantum gating for the estimation of task relevance, which is motivated by the distribution gap between the pretrained model and downstream tasks. The $\mathcal{A}_1$ effectively bridges gap and captures generalizable base semantics that are more suitable for subsequent quantum encoding. We observe that using multiple adapters for feature extraction introduces redundant computation while yielding only marginal performance differences. In Tab.~\ref{tab:adapter_diff}, we compare the performance of different settings. The results demonstrate that $f_{\mathrm{ViT}}(x; \mathcal{A}_1)$ provides a favorable trade-off between performance and efficiency.
\section{Conclusion}
\label{sec:con}
In this work, we addressed the long-standing challenge of modeling task interaction in CIL, where explicit task boundaries disappear during inference and feature subspaces across tasks tend to become entangled. To overcome these limitations, we introduced Quantum-Gated Knowledge Distillation (QKD), a novel framework that integrates quantum machine learning with PTM-based CIL. By encoding task-level representations into quantum states and exploiting quantum interference to estimate sample–task correlation, QKD provides an explicit, learnable mechanism for routing both training signals and inference paths across multiple adapters. The resulting correlation-weighted feature distillation enables new adapters to selectively absorb informative knowledge from relevant tasks while avoiding interference from unrelated ones. Extensive experiments on five benchmarks demonstrate that QKD consistently outperforms state-of-the-art methods, highlighting its robustness, generalization ability, and effectiveness. 

\section*{Acknowledgments}
This work is supported in part by the National Natural Science Foundation of China (No. 62101061) and in part by the Cooperation Agreement Project of Kunpeng-Shengteng Science, Education, Innovation and Incubation Center, Beijing University of Posts and Telecommunications.
{
    \small
    \bibliographystyle{ieeenat_fullname}
    \bibliography{main}
}


\end{document}